\documentclass[lettersize,journal]{IEEEtran}

\usepackage{amsmath,amsfonts,amssymb}
\usepackage{mathtools}
\usepackage{bm}
\usepackage{mathptmx} 
\DeclareMathAlphabet{\mathcal}{OMS}{cmsy}{m}{n}
\DeclareSymbolFont{largesymbols}{OMX}{cmex}{m}{n}

\usepackage{graphicx}
\usepackage{epsfig}
\usepackage{epstopdf}
\usepackage{float}  
\usepackage{stfloats}
\usepackage{array}
\usepackage{booktabs}
\usepackage{multirow}
\usepackage{longtable}
\usepackage{supertabular}

\usepackage{subfig}

\usepackage[linesnumbered,ruled,vlined]{algorithm2e}
\usepackage{algorithmic}
\usepackage{xcolor}
\usepackage{soul}
\usepackage{color}
\usepackage{textcomp}
\usepackage{url}
\usepackage{verbatim}
\usepackage{cite}
\usepackage{CJKutf8}
\usepackage{changepage}
\usepackage[mathscr]{euscript}
\usepackage{cancel}
\usepackage{balance}
\usepackage{diagbox}
\usepackage{pifont}
\usepackage{fixltx2e}
\usepackage{threeparttable}
\usepackage{slashbox}
\usepackage{makecell}
\usepackage{tikz}
\usepackage[linkcolor=black,citecolor=black,urlcolor=black,colorlinks=true]{hyperref}

\let\oldtwocolumn\twocolumn
\renewcommand\twocolumn[1][]{%
    \oldtwocolumn[{#1}{
    \begin{center}
           \includegraphics[width=0.9\textwidth]{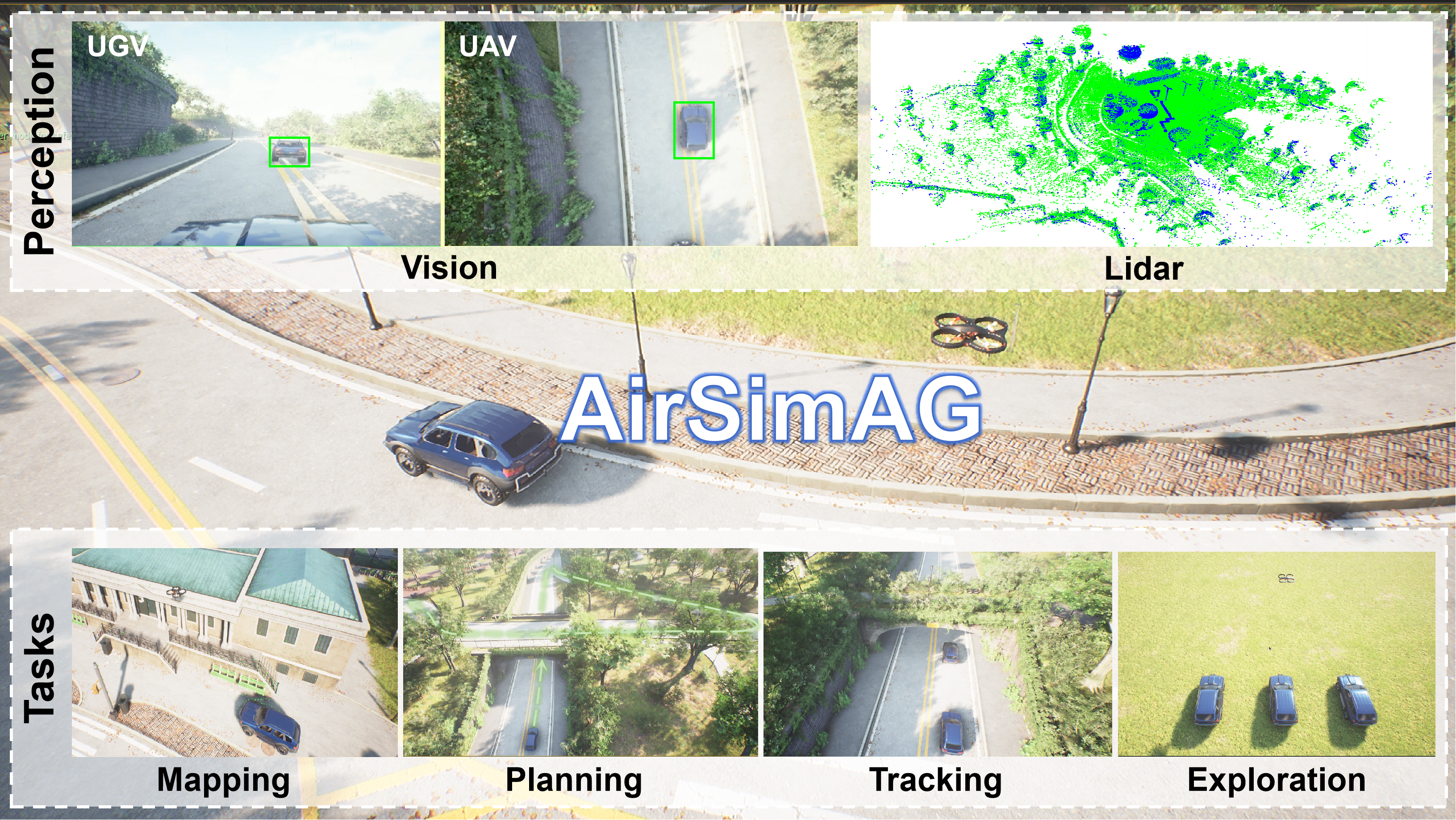}
           \captionof{figure}{\label{fig:top}  Overview of AirSimAG. The platform supports air–ground collaborative perception and closed-loop simulation, with representative cooperative experiments designed for validation. }
           \label{fig:show_result}
        \end{center}
    }]
}

\begin{document}
\title{AirSimAG: A High-Fidelity Simulation Platform for Air-Ground Collaborative Robotics}

\author{Yangjie Cui, Xin Dong, Boyang Gao, Jinwu Xiang, Daochun Li, Zhan Tu*
\thanks{
Yangjie Cui, Jinwu Xiang, and Daochun Li are with the School of Aeronautic Science and Engineering, Beihang University, Beijing 100191, China.
Xin Dong and Boyang Gao are with the Hangzhou International Innovation Institute, Beihang University, Hangzhou 311115, China. 
Zhan Tu is with the Institution of Unmanned System, Beihang University, Beijing 100191, China. Corresponding author: Zhan Tu (zhantu@buaa.edu.cn). }}

\maketitle

\setcounter{figure}{1}

\begin{abstract}
As spatial intelligence continues to evolve, heterogeneous multi-agent systems—particularly the collaboration between Unmanned Aerial Vehicles (UAVs) and Unmanned Ground Vehicles (UGVs)—have demonstrated strong potential in complex applications such as search and rescue, urban surveillance, and environmental monitoring. However, existing simulation platforms are primarily designed for single-agent dynamics and lack dedicated frameworks for interactive air–ground collaborative simulation.
In this paper, we present AirsimAG, a high-fidelity air–ground collaborative simulation platform built upon an extensively customized AirSim framework. The platform enables synchronized multi-agent simulation and supports heterogeneous sensing and control interfaces for UAV–UGV systems.
To demonstrate its capabilities, we design a set of representative air–ground collaborative tasks, including mapping, planning, tracking, formation, and exploration. We further provide quantitative analyses based on these tasks to illustrate the platform’s effectiveness in supporting multi-agent coordination and cross-modal data consistency. The AirsimAG simulation platform is publicly available at: https://github.com/BIULab-BUAA/AirSimAG. 
\end{abstract}


\section{Introduction}
\label{sec:intro}
\IEEEPARstart{T}{he} rapid development of spatial intelligence is shifting autonomous systems from single-agent autonomy to heterogeneous multi-agent collaboration. Among these paradigms, Air–Ground Collaborative Systems (AGCS), which combine Unmanned Aerial Vehicles (UAVs) with Unmanned Ground Vehicles (UGVs), have attracted significant attention \cite{stronger_together_2022}. UAVs provide global perception from high altitude, while UGVs enable precise interaction with the environment. This complementarity makes AGCS suitable for tasks such as urban search and rescue, large-scale mapping, and long-term surveillance in complex environments.

Despite these advantages, the development of air–ground collaborative algorithms remains challenging due to the lack of dedicated datasets and simulation tools. Most existing datasets focus on either aerial or ground viewpoints \cite{uav123, DSEC-Dataset}. Such datasets cannot capture cross-view spatial–temporal correlations required for air-ground cooperative perception. Some recent works \cite{griffin_2025} attempt to generate air–ground collaborative perception datasets by integrating multiple simulation platforms. However, these solutions are often ad hoc and lack a unified design for air–ground collaboration. In addition, real-world data collection is costly and risky, especially during early-stage validation.

Simulation platforms have therefore become an important alternative for data generation and system validation. However, current simulators exhibit notable limitations in supporting heterogeneous air–ground collaboration. Popular platforms such as AirSim \cite{airsim_2017}, XTDrone \cite{XTDrone_2020}, and CARLA \cite{CARLA_2017} are primarily designed for either aerial or ground systems, and provide limited support for tightly coupled multi-agent interaction. 
Some general-purpose simulators, such as Gazebo\cite{gazebo_2004} and Isaac-based frameworks\cite{isaaclab_2025}, can support air–ground scenarios through manual configuration. However, Gazebo provides limited support for multi-agent air–ground collaborative perception. Isaac-based frameworks often require substantial development effort. They lack unified modules for independent dynamics modeling and perception configuration. This leads to high setup complexity and limits their usability for rapid experimentation. In particular, challenges remain in achieving fine-grained temporal synchronization across heterogeneous agents, aligning multi-modal data from different viewpoints, and representing complex collaborative behaviors in dynamic environments.

To address these limitations, we present \textit{AirSimAG}, a high-fidelity simulation framework for air–ground collaborative systems. The framework is built on an extended AirSim core. It supports synchronized multi-agent simulation through unified sensing and control interfaces. The system enables high-frequency multi-modal data acquisition with consistent timestamps. It also includes an interactive mission planner for flexible scenario construction.
To evaluate the proposed framework, we design several representative tasks, including mapping, planning, tracking, and formation control. These tasks are used to assess coordination performance and cross-view perception consistency. The experiments also demonstrate scalability in complex environments.
Based on these tasks, we provide quantitative analyses to illustrate the platform’s ability to support multi-agent coordination and cross-view data consistency in complex environments.

The main contributions of this paper are summarized as follows:
\begin{itemize}
    \item An extended simulation platform for air–ground systems, termed AirSimAG, is developed. The platform enables synchronized operation among heterogeneous agents and supports high-frequency multi-modal data acquisition in multi-agent scenarios.

    \item A set of representative air–ground collaborative tasks, including mapping, tracking, navigation, and formation, has been designed to demonstrate the functionality and flexibility of the proposed platform in diverse scenarios.

    \item Quantitative analyses based on the designed tasks are provided, illustrating the capability of the platform to support coordinated behaviors and cross-view perception in heterogeneous multi-agent systems.
\end{itemize}

\section{Related Work}
\label{sec:related_work}

\subsection{Simulation Platform and Dataset}

Simulation platforms play a critical role in the development and evaluation of autonomous systems, offering safe, scalable, and cost-effective environments for algorithm validation and data generation. By enabling controlled experimentation and access to labeled data, these platforms facilitate the transition from algorithm design to real-world deployment. 

\textbf{General-Purpose Robotics Simulators.} 
Early robotic simulation platforms such as Gazebo \cite{gazebo_2004} provide mature physics engines and tight integration with Robot Operating System (ROS), making them widely adopted in multi-robot systems. However, their limited visual realism constrains their applicability in perception-driven tasks. To address this limitation, AirSim \cite{airsim_2017} introduced a photorealistic simulation framework based on Unreal Engine, supporting UAVs or ground vehicles with improved rendering fidelity. More recently, GPU-accelerated platforms such as NVIDIA Isaac Lab \cite{isaaclab_2025} have enabled large-scale parallel simulation for robot learning, although achieving high-fidelity outdoor environments often incurs substantial computational overhead.

\textbf{Domain-Specific Extensions of AirSim.} 
To better support specialized applications, several extensions of AirSim have been developed. AirSim-W \cite{airsim-w_2018} adapts the platform for wildlife monitoring scenarios, while COSYS-AIRSIM \cite{COSYS-AIRSIM_2023} and ASVSim \cite{ASVSim_2025} extend its capabilities to industrial sensing and autonomous surface vehicles, respectively. AirSim360 \cite{airsim360_2025} introduces panoramic perception for UAV platforms, enabling broader environmental awareness. In parallel, platforms such as XTDrone \cite{XTDrone_2020} focus on multi-rotor control and swarm simulation, whereas UavNetSim-v1 \cite{UavNetSim_2025} emphasizes communication-aware multi-UAV systems.

\textbf{Air-Ground Collaborative Datasets and Gaps.} 
Despite these developments, resources explicitly designed for air–ground collaboration remain limited. Existing datasets are typically constrained to single-agent perspectives, such as ground vehicle datasets \cite{kitti} or aerial datasets \cite{VisDrone}, which are insufficient for studying cross-view perception and coordinated behaviors. Recent efforts, such as the Griffin \cite{griffin_2025} dataset, begin to explore air-ground cooperative perception; however, they are often restricted by fixed sensor configurations and limited scenario diversity.

Existing simulation platforms and datasets have significantly advanced the development of autonomous systems. However, they remain insufficient for air–ground collaborative research. General-purpose simulators provide strong physics support but lack realistic perception or unified multi-agent coordination. Domain-specific extensions improve certain capabilities but are typically tailored to single modalities or specific tasks. Meanwhile, existing datasets are mostly limited to single-agent perspectives, with only a few recent attempts exploring air–ground cooperation under constrained settings. Overall, there is still a lack of a unified, high-fidelity, and extensible simulation framework that supports synchronized air–ground interaction, cross-view perception, and scalable multi-agent collaboration.

\subsection{Air-ground Collaborative Task}

The collaboration between Unmanned Air Vehicles (UAVs) and Unmanned Ground Vehicles (UGVs) has become an important paradigm for a wide range of autonomous applications, including environmental exploration, infrastructure inspection, and search and rescue. By combining complementary sensing perspectives and mobility characteristics, air–ground systems can improve both efficiency and robustness in complex environments.

\textbf{Collaborative Localization and Mapping.} 
Accurate spatial perception is a fundamental requirement for air–ground systems operating in challenging environments. Prior work has investigated collaborative localization in GNSS-denied scenarios \cite{location_nognss_2022} and industrial environments \cite{zhang2018intelligent}, as well as specialized methods leveraging LiDAR-based representations for forested areas \cite{localisation_forest_2023}. In terms of mapping, existing studies have explored risk-aware terrain mapping for off-road navigation \cite{mapping1_2023} and semantic active mapping strategies for efficient exploration \cite{same_2024}. Despite these advances, achieving consistent and high-fidelity mapping across heterogeneous viewpoints remains challenging, particularly due to differences in sensing modalities and viewpoints between air and ground platforms.

\textbf{Cooperative Exploration and Planning.} 
The integration of UAV global perception with UGV local navigation has enabled a variety of cooperative exploration and planning strategies. For example, aerial-assisted exploration approaches improve coverage efficiency in large-scale unknown environments \cite{aage_2025}, while collaborative planning methods incorporate shared spatial representations to guide ground navigation \cite{navigation_2024,yu2014cooperative}. Nevertheless, coordinating decision-making across heterogeneous agents remains nontrivial, especially in dynamic environments where timely information exchange and consistent environmental understanding are required.

\textbf{Collaborative Detection and Tracking.} 
Air–ground collaboration has also been widely explored in detection and tracking tasks, where aerial viewpoints can alleviate occlusion and provide global context. Prior work includes hierarchical tracking frameworks using aerial perspectives \cite{air_ground_tracking_2025} and cooperative vision-based localization methods \cite{minaeian2015vision}. However, practical deployment is often affected by cross-view inconsistencies, including viewpoint misalignment and temporal asynchrony, which complicate data association and robust feature matching.

Although substantial progress has been made across these task domains, a common challenge lies in the lack of unified experimental platforms for systematic evaluation. Existing studies are often conducted under task-specific settings or simplified environments, making it difficult to analyze multi-stage interactions and cross-view consistency in a controlled manner. In particular, limitations in temporal synchronization, multi-modal data alignment, and interactive scenario design restrict the reproducibility and comparability of air–ground collaborative research.

To address these challenges, this work presents AirSimAG, a simulation platform designed to support synchronized multi-agent operation and flexible construction of representative air–ground collaborative tasks, enabling systematic analysis of coordination behaviors and cross-view perception in complex environments.

\section{AirsimAG Platform}
\label{sec:platform}
To address the limitations of the original AirSim in heterogeneous multi-agent scenarios, an extended platform, termed AirSimAG, is developed. The proposed platform is designed to overcome architectural constraints in AirSim that hinder the simultaneous operation of UAVs and UGVs.

\subsection{Architecture}
The design of AirSimAG is motivated by the architectural limitations of the original AirSim framework, particularly its reliance on the \textit{SimMode} base class, which centrally manages vehicles, sensors, and world states. While this design is effective for single-agent simulation, it becomes restrictive in multi-agent settings involving heterogeneous platforms. Specifically, the tight coupling of sensor management, vehicle control, and environment access within \textit{SimMode} introduces several limitations, which hinder the development of synchronized and interactive air–ground collaborative scenarios.

\begin{figure}
    \centering
    \includegraphics[width=1.0\linewidth]{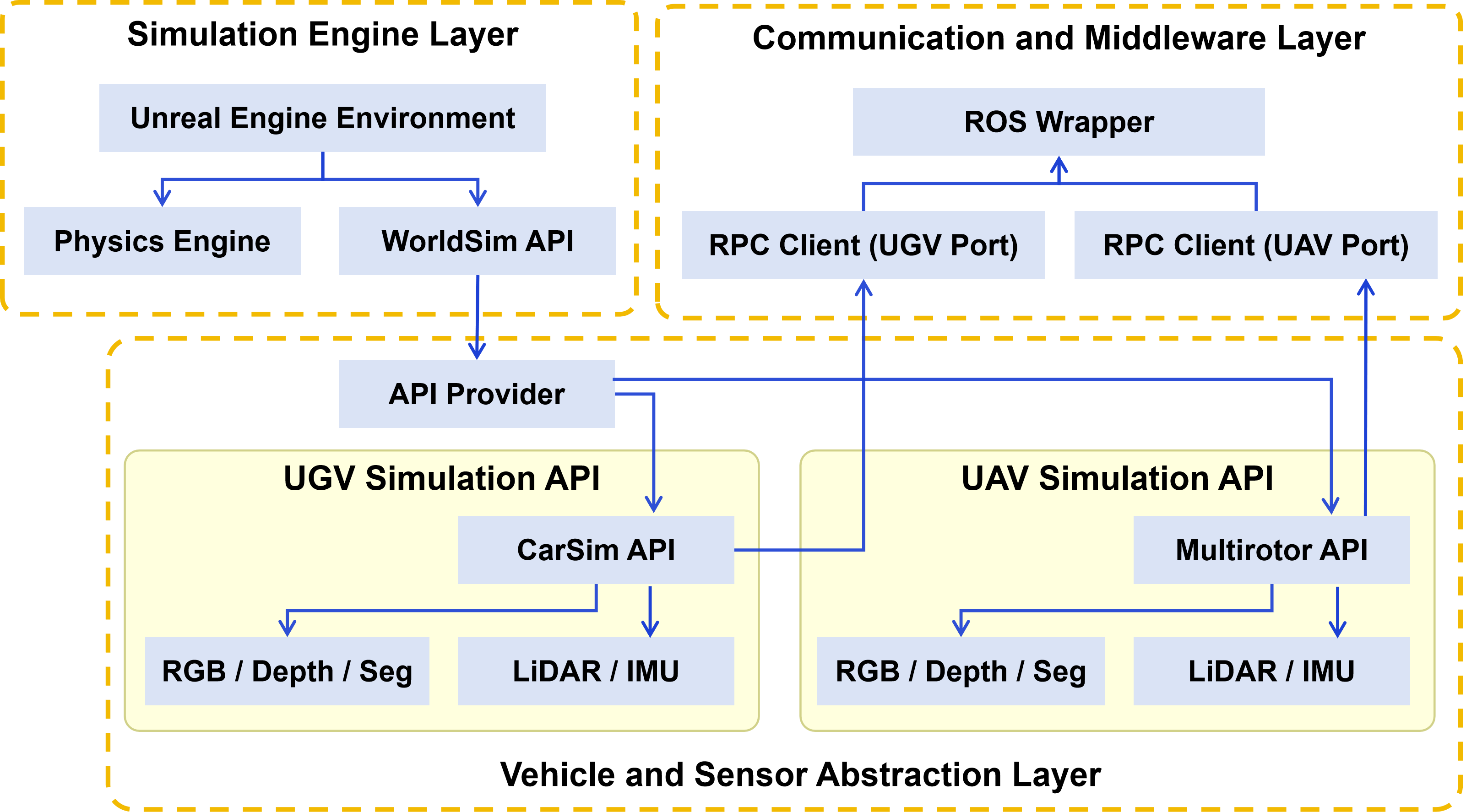}
    \caption{Overall system architecture of AirSimAG. }
    \label{fig:architecture}
\end{figure}

To overcome these limitations, AirSimAG adopts a decoupled system architecture that separates vehicle management, sensor interfaces, and communication modules. This design enables scalable and stable simulation of multiple heterogeneous agents equipped with diverse sensing modalities. At the same time, compatibility with the existing AirSim ecosystem is preserved, facilitating integration with established tools and workflows. The overall system architecture of AirSimAG is illustrated in Fig.~\ref{fig:architecture}.

\textbf{Simulation Engine Layer.}
The simulation engine layer of AirSimAG is built on Unreal Engine, which provides high-fidelity rendering, physics simulation, and environment modeling. In the original AirSim framework, access to world states, sensors, and vehicles is tightly coupled within the \textit{SimMode} base class. This design creates a single control entry for the entire simulation. While effective for single-agent scenarios, it limits flexibility in multi-agent settings, especially when heterogeneous platforms operate simultaneously.

To overcome this limitation, AirSimAG decouples the WorldSimApi interface from \textit{SimMode}. Instead of relying on a single shared instance, an API-based mechanism is introduced. Each vehicle registers an independent simulation API instance. Sensor data and camera streams are retrieved directly from the corresponding vehicle API. This design enables the concurrent operation of multiple UAVs and UGVs within the same environment, while preserving independent control and perception.

\textbf{Vehicle and Sensor Abstraction Layer.}
This layer provides a unified abstraction for heterogeneous vehicles and their sensors. Each vehicle is associated with a dedicated simulation API, which manages vehicle states, sensor configurations, and coordinate transformations. The API mechanism dynamically resolves vehicle-specific interfaces based on vehicle identifiers. As a result, sensor operations, such as image capture and LiDAR acquisition, are routed to the correct vehicle without a global controller.

The framework supports multiple sensor modalities, including RGB, depth, and semantic cameras, as well as LiDAR and inertial sensors. In addition, AirSimAG implements a global coordinate transformation module. When \textit{SimMode} is bypassed, global NED transformations are obtained directly from vehicle APIs. This ensures consistent spatial alignment across all agents and sensors.

\textbf{Communication and Middleware Layer.}
The communication layer provides a robust interface between AirSimAG and external systems, such as ROS-based autonomy stacks. In the original AirSim implementation, a single RPC client handles all communication. This design can lead to conflicts in multi-vehicle scenarios.
AirSimAG adopts a multi-client communication architecture. Each vehicle type is assigned an independent RPC client with a dedicated communication port. In the current implementation, multirotor UAVs and UGVs use separate clients. This separation prevents data conflicts and ensures the correct routing of sensor requests.

On the ROS side, the AirSim ROS wrapper is extended to support dynamic request routing. Sensor requests are dispatched based on vehicle type. UAV data are handled by the multirotor client, while UGV data are handled by the car client. This mechanism guarantees consistency between sensor data and simulation instances, and avoids empty or mismatched responses.
The communication layer enables reliable integration with external frameworks. It supports both control commands and high-bandwidth sensor streams, while maintaining scalability and independence in heterogeneous multi-agent simulation.

\subsection{Data Collection}

A key objective of AirSimAG is to support reliable air-ground perception and real-time interaction. To achieve this, the platform implements a structured data collection pipeline that ensures temporal consistency, spatial alignment, and synchronized sensing across heterogeneous agents.

\textbf{Time Synchronization.}
Accurate temporal alignment is critical for sensor fusion and coordinated multi-agent operation. AirSimAG maintains a unified simulation clock within the Unreal Engine environment. All sensor data are generated under the same simulation timestep. Sensor requests from external interfaces, such as ROS, are processed within synchronized update cycles. This design ensures that data streams from different agents correspond to the same simulation state. As a result, the platform supports consistent multi-view reconstruction and cooperative perception.

\textbf{Coordinate System Alignment.}
AirSimAG adopts a unified global coordinate system based on the North-East-Down (NED) convention. To ensure consistent spatial references, the platform provides direct access to global transformations through vehicle simulation APIs. Each vehicle retrieves its pose in the global frame without relying on a centralized controller. This mechanism guarantees that all vehicles, sensors, and environmental elements share a consistent spatial reference. It is essential for tasks that require precise alignment between aerial observations, ground states, and scene geometry.

\textbf{Perception Data Generation.}
Based on the synchronization mechanisms above, AirSimAG supports multi-modal perception data generation from heterogeneous agents. Each vehicle can acquire multiple sensor streams in parallel. These include RGB images, depth maps, semantic segmentation, and LiDAR point clouds. Data can be collected from multiple viewpoints, enabling comprehensive coverage from both aerial and ground perspectives. Sensor parameters, such as camera intrinsics, LiDAR settings, and sensor placement, are fully configurable. This flexibility allows the platform to adapt to diverse experimental requirements.

By integrating synchronized sensing, consistent spatial alignment, and scalable multi-agent operation, AirSimAG enables reproducible dataset generation. It also provides a controlled environment for evaluating algorithms in air–ground collaborative perception and control.

\section{Cooperative Tasks and Experiments}
\label{sec:exp}

\begin{figure}
    \centering
    \includegraphics[width=1.0\linewidth]{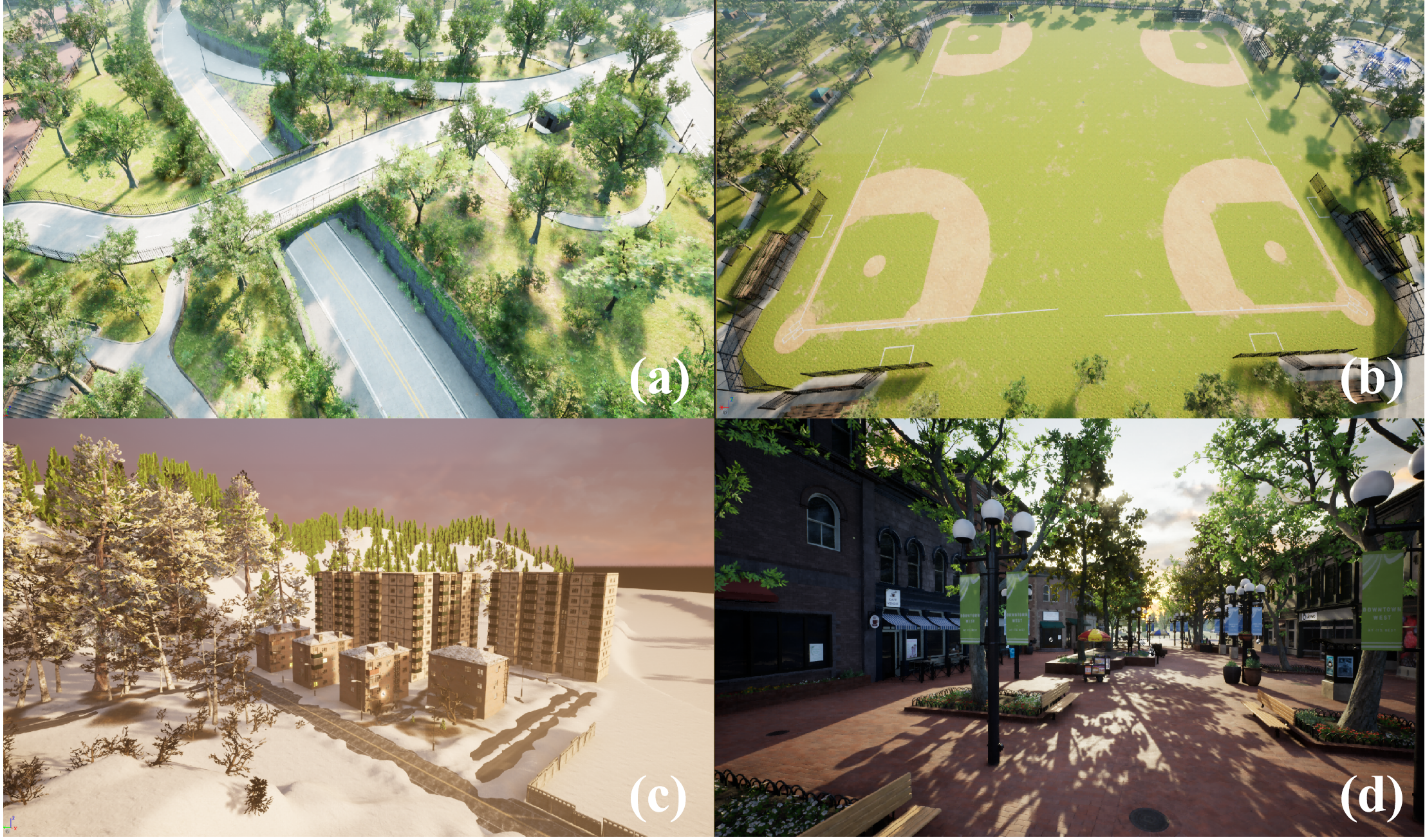}
    \caption{Different Maps}
    \label{fig:maps}
\end{figure}

To demonstrate the capability of AirSimAG in heterogeneous air–ground scenarios, we design a set of representative cooperative tasks involving both UAVs and UGVs. These tasks evaluate key aspects of air–ground collaboration, including perception, planning, perception–action integration, and scalable multi-agent coordination. AirSimAG supports multiple embedded environments, and the experiments are primarily conducted in the scene shown in Fig.\ref{fig:maps}.
Specifically, four tasks are implemented in the AirSimAG environment:
\begin{itemize}
    \item Cooperative mapping using multi-platform LiDAR sensing;
    \item Aerial-assisted ground vehicle navigation for perception-driven planning;
    \item Cooperative multi-view target tracking with perception–planning integration;
    \item Scalable multi-agent formation and coordination for system-level evaluation.
\end{itemize}

These tasks collectively assess the ability of AirSimAG to support multi-sensor fusion, perception-driven decision making, coordinated tracking, and scalable deployment. The following subsections describe each task and present the corresponding results.

\begin{figure*}
    \centering
    \includegraphics[width=0.8\linewidth]{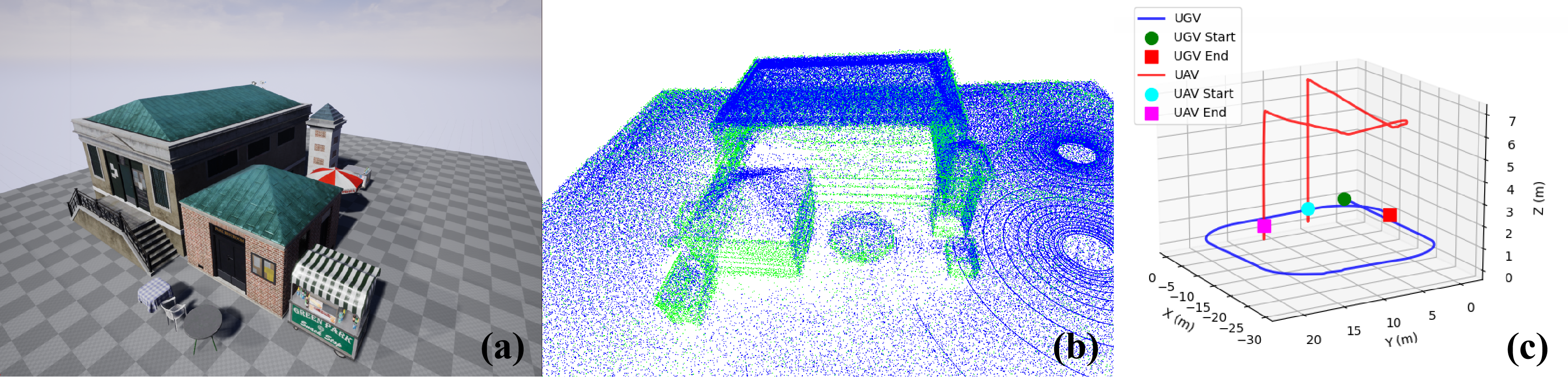}
    \caption{Air–ground collaborative mapping. (a) Mapping scene. (b) Fused point cloud map: UAV points in blue, UGV points in green. (c) UAV and UGV trajectories during the mapping task.}
    \label{fig:task_mapping}
\end{figure*}

\begin{figure*}
    \centering
    \includegraphics[width=0.8\linewidth]{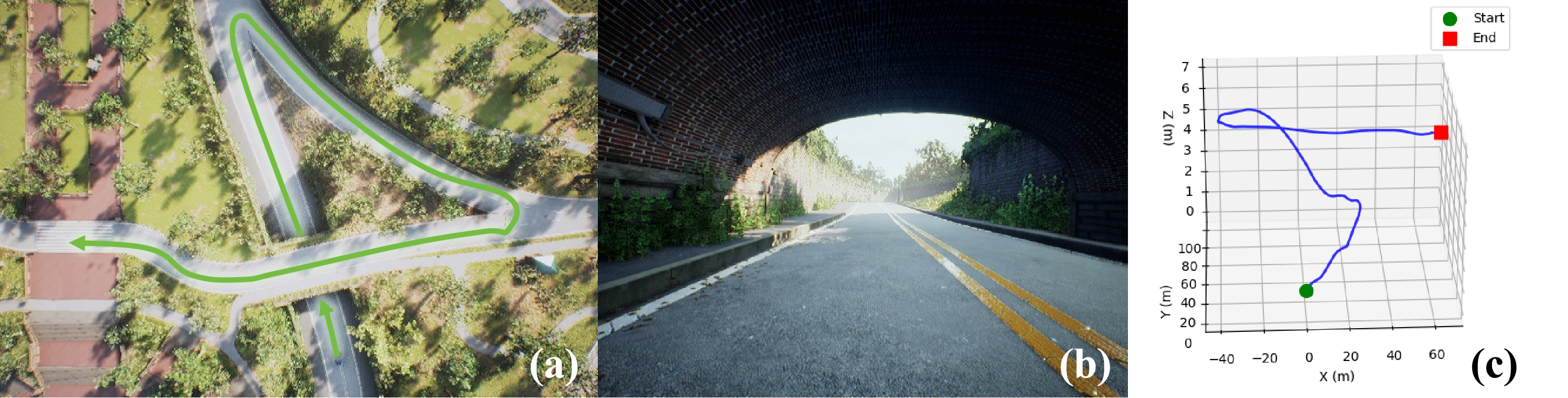}
    \caption{Air–ground collaborative planning. (a) UGV planned path from UAV’s first-person view. (b) UGV first-person view during planning. (c) Executed UGV trajectory. }
    \label{fig:task_planning}
\end{figure*}

\subsection{Cooperative Mapping}
\label{subsec:mapping}
Accurate environment mapping is essential for autonomous navigation and situational awareness. UGVs provide dense local measurements but suffer from occlusions and limited sensing range. UAVs offer a broader field of view from elevated positions but provide less detailed ground observations.

To leverage these complementary properties, we implement a cooperative mapping task that fuses LiDAR data from both platforms. The UAV performs aerial scanning to capture a global structure of the environment. The UGV collects dense point clouds during ground traversal. All measurements are transformed into a unified global frame using synchronized poses provided by AirSimAG. The Iterative Closest Point (ICP) method \cite{ICP} is then applied to align aerial and ground point clouds.

The mapping results are shown in Fig.~\ref{fig:task_mapping}. The fused map exhibits improved completeness compared with single-platform reconstruction. Quantitative results are summarized in Table~\ref{tb:mapping}. The UGV completes a trajectory of 98.0 m in 42.4 s, with an average speed of 2.3 m/s. The UAV operates for 54.5 s with a trajectory length of 68.8 m and an average speed of 1.3 m/s. These results highlight the complementary roles of the two agents. The UGV enables efficient exploration, while the UAV provides stable global coverage.

For registration accuracy, ICP estimates a translation of $[-0.176, -0.097, 0.195]$ m with a root mean square error (RMSE) of 0.273 m. The low RMSE indicates accurate alignment between aerial and ground LiDAR data, despite differences in viewpoint and sensing conditions. These results demonstrate that AirSimAG supports reliable cooperative perception and provides a solid basis for downstream tasks such as planning and tracking.

\begin{table}[]
\centering
\caption{Kinematic statistics of the UAV and UGV during mapping, and point cloud registration performance using ICP.}
\begin{tabular}{c|c|c}
\hline
\multirow{3}{*}{UGV}                & Duration(s)         & 42.4                        \\
                                    & Total Length (m)    & 98.0                        \\
                                    & Averge Speed (m/s)  & 2.3                         \\ \hline
\multirow{3}{*}{UAV}                & Duration(s)         & 54.5                        \\
                                    & Total Length (m)    & 68.8                        \\
                                    & Averge Speed (m/s)  & 1.3                         \\ \hline
\multirow{2}{*}{Cloud Points (ICP)} & Est Translation (m) & {[}-0.176, -0.097, 0.195{]} \\
                                    & RMSE (m)            & 0.273                       \\ \hline
\end{tabular}
\label{tb:mapping}
\end{table}

\subsection{Aerial-assisted Ground Vehicle Navigation}
\label{subsec:planning}
Efficient navigation in complex environments requires both global awareness and local obstacle avoidance. UGVs rely on onboard sensors for local perception, but their limited field of view restricts global path optimality.

To address this limitation, we implement a cooperative planning task in which a UAV assists UGV navigation. The UAV captures top-down images to construct a global occupancy map. Based on this map, a high-level path is generated using a planning algorithm. The planned trajectory is transmitted to the UGV through the AirSimAG communication interface. During execution, the UGV follows the global path while performing local obstacle avoidance.

An example is shown in Fig.~\ref{fig:task_planning}. From a high-altitude viewpoint (approximately 85 m), the UAV provides global situational awareness. This enables the UGV to execute complex maneuvers, such as navigating under a bridge and ascending onto it. The vehicle reaches the target area efficiently. Quantitative results are summarized in Table~\ref{fig:task_planning}. The UGV completes a trajectory of 309.7 m in 86.0 s, with an average speed of 3.6 m/s. The UGV exhibits full 3D motion, with variations along the vertical axis caused by elevation changes during traversal, such as slope climbing (X: [-39.6, 65.2] m, Y: [0.0, 102.5] m, Z: [0.0, 6.9] m). The UAV maintains continuous aerial coverage throughout the task. These results demonstrate effective coordination between global planning and local execution, enabled by air–ground collaboration.

\begin{figure*}[htbp!]
    \centering
    \includegraphics[width=0.8\linewidth]{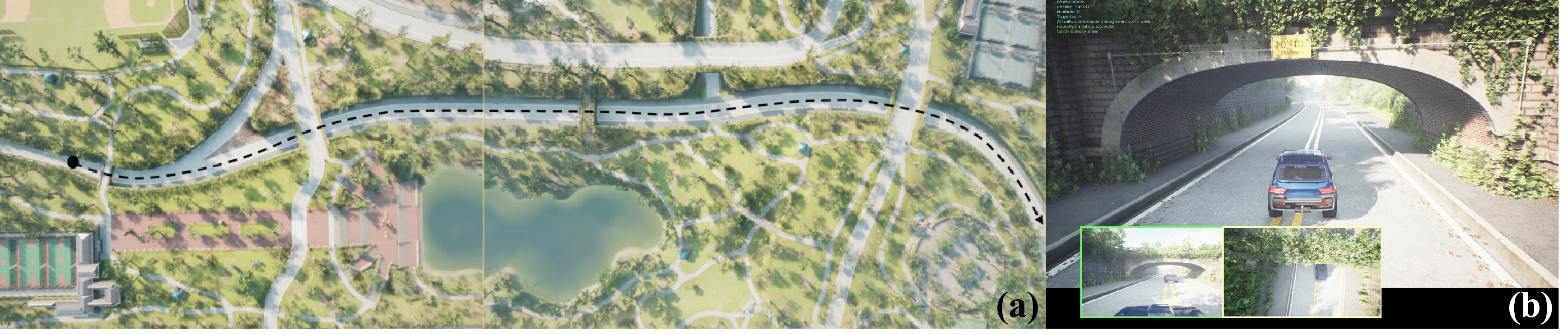}
    \caption{Air–ground collaborative tracking. (a) Tracking trajectories and map. (b) Unreal Engine scene showing the third-person view, UAV first-person view, and UGV first-person view. }
    \label{fig:task_tracking0}
\end{figure*}

\begin{figure*}[htbp!]
    \centering
    \includegraphics[width=0.9\linewidth]{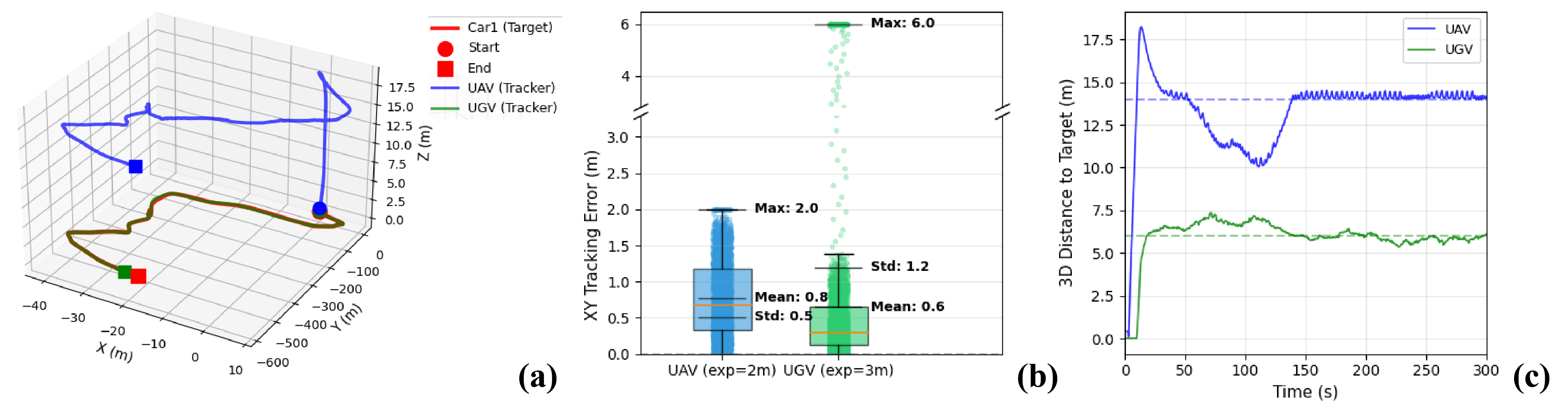}
    \caption{Air–ground collaborative tracking. (a) Trajectories of UAV, UGV, and target. (b) Tracking errors on the X–Y plane, including maximum, mean, and variance. (c) Time-varying 3D distances from UAV and UGV to the target. The dotted line represents the expected distance (14.0 m for UAV, and 6.0m for UGV). }
    \label{fig:task_tracking1}
\end{figure*}

\begin{table}[htbp!]
\centering
\caption{Kinematic statistics of the UAV and UGV during planning. }
\begin{tabular}{cc}
\hline
\textbf{Metric} & \textbf{Data} \\ \hline
Duration (s)        & 86.0              \\ 
Total Length (m)    & 309.7             \\ 
Average Speed (m/s) & 3.6               \\ 
X Range (m)         & {[}-39.6, 65.2{]} \\ 
Y Range (m)         & {[}0.0, 102.5{]}  \\ 
Z Range (m)         & {[}0.0, 6.9{]}    \\ \hline
\end{tabular}
\end{table}

\begin{figure}[htbp!]
    \centering
    \includegraphics[width=1.0\linewidth]{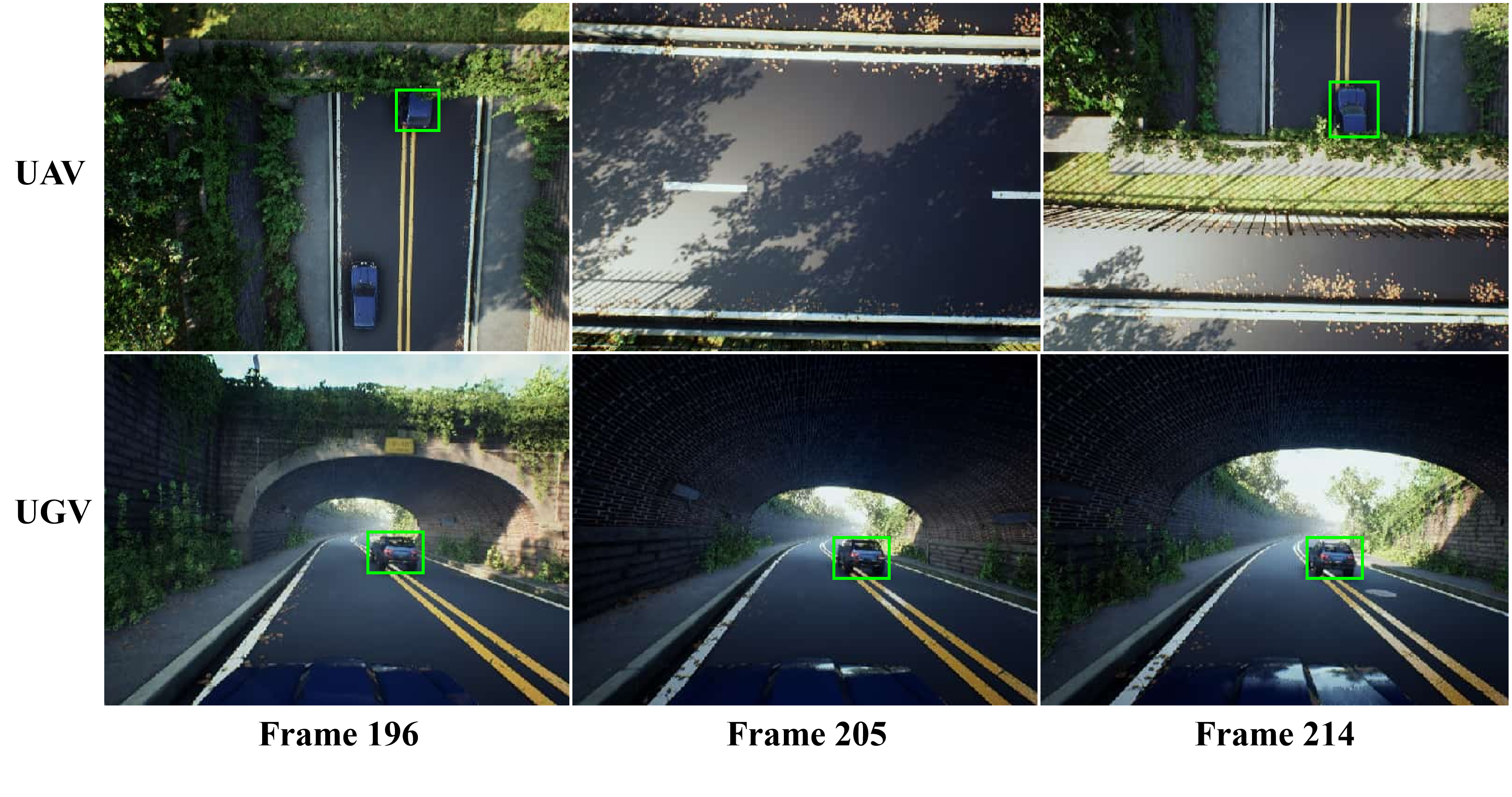}
    \caption{First view of UAV and UGV (tracker) when the target passes through the aperture of the bridge.}
    \label{fig:task_tracking2}
\end{figure}

\begin{figure}[htbp!]
    \centering
    \includegraphics[width=0.8\linewidth]{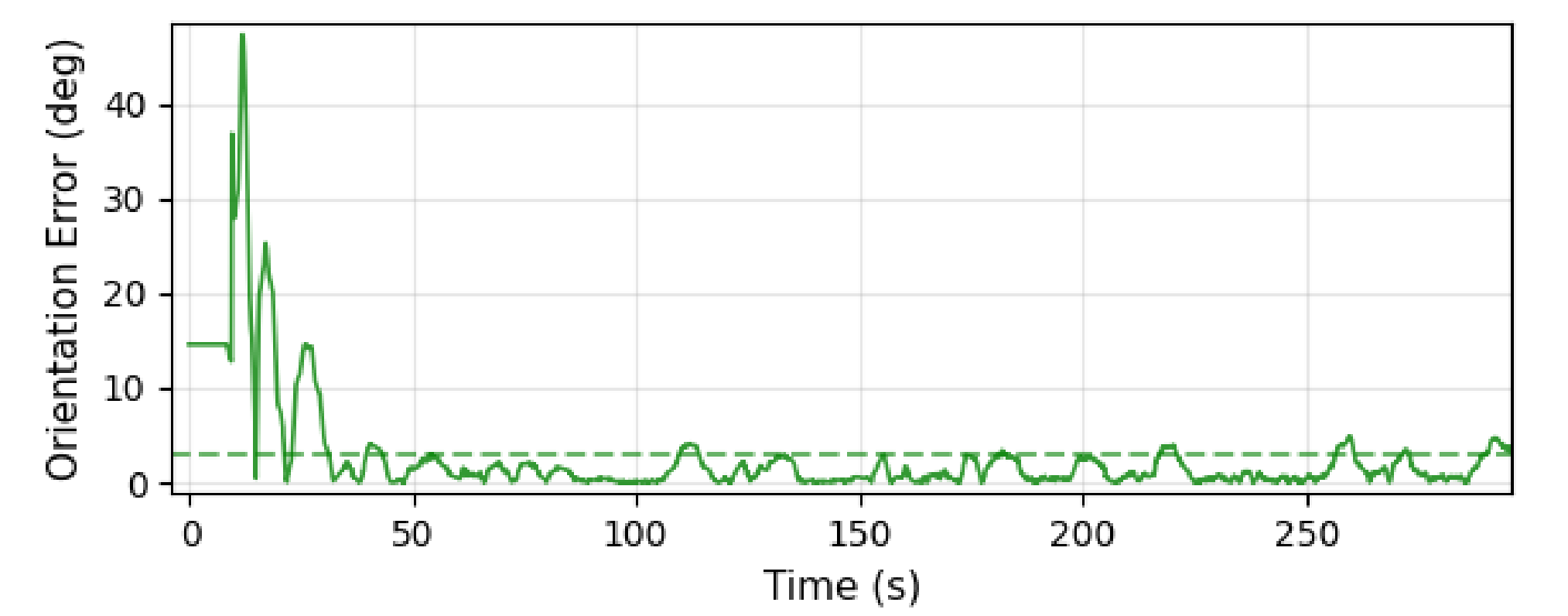}
    \caption{Orientation Error of UGV (tracker) in the tracking process. The dotted line represents the average error (2.91 deg).}
    \label{fig:task_tracking3}
\end{figure}

\subsection{Cooperative Multi-view Target Tracking}
\label{subsec:tracking}

Target tracking is a key capability in applications such as surveillance and search and rescue. Single-agent systems often fail under occlusions or limited viewpoints.

To overcome these limitations, we design a cooperative tracking task involving UAVs and UGVs. The UAV observes the target from an aerial perspective and provides global context. The UGV tracks the target at close range using onboard sensors. Observations from both agents are fused to maintain continuous tracking.

The tracking scenario is illustrated in Fig.~\ref{fig:task_tracking0}. When the target is occluded from the UAV, the UGV continues tracking and provides position updates. First-person views from both agents are shown in Fig.~\ref{fig:task_tracking2}, where the target passes under a bridge.
The task is evaluated over a 600 m trajectory with a duration of approximately 300 s. Quantitative results are shown in Figs.~\ref{fig:task_tracking1} and \ref{fig:task_tracking3}. The UAV achieves a mean tracking error of 0.8 m, with a variance of 0.5 m and a maximum error of 2.0 m. The UGV achieves a mean error of 0.6 m, with a variance of 1.2 m and a maximum error of 6.0 m. The 3D distance curves show that both agents reach a stable tracking regime after approximately 150 s. Early errors correspond to the initial coordination phase.
We further evaluate orientation stability using the UGV yaw error. The error decreases rapidly and remains below an average of 2.49 $deg$ after 50 $s$. This indicates stable heading control during most of the task. These results demonstrate that cooperative perception improves robustness to occlusion and viewpoint limitations.

\begin{figure}[tbp!]
    \centering
    \includegraphics[width=0.8\linewidth]{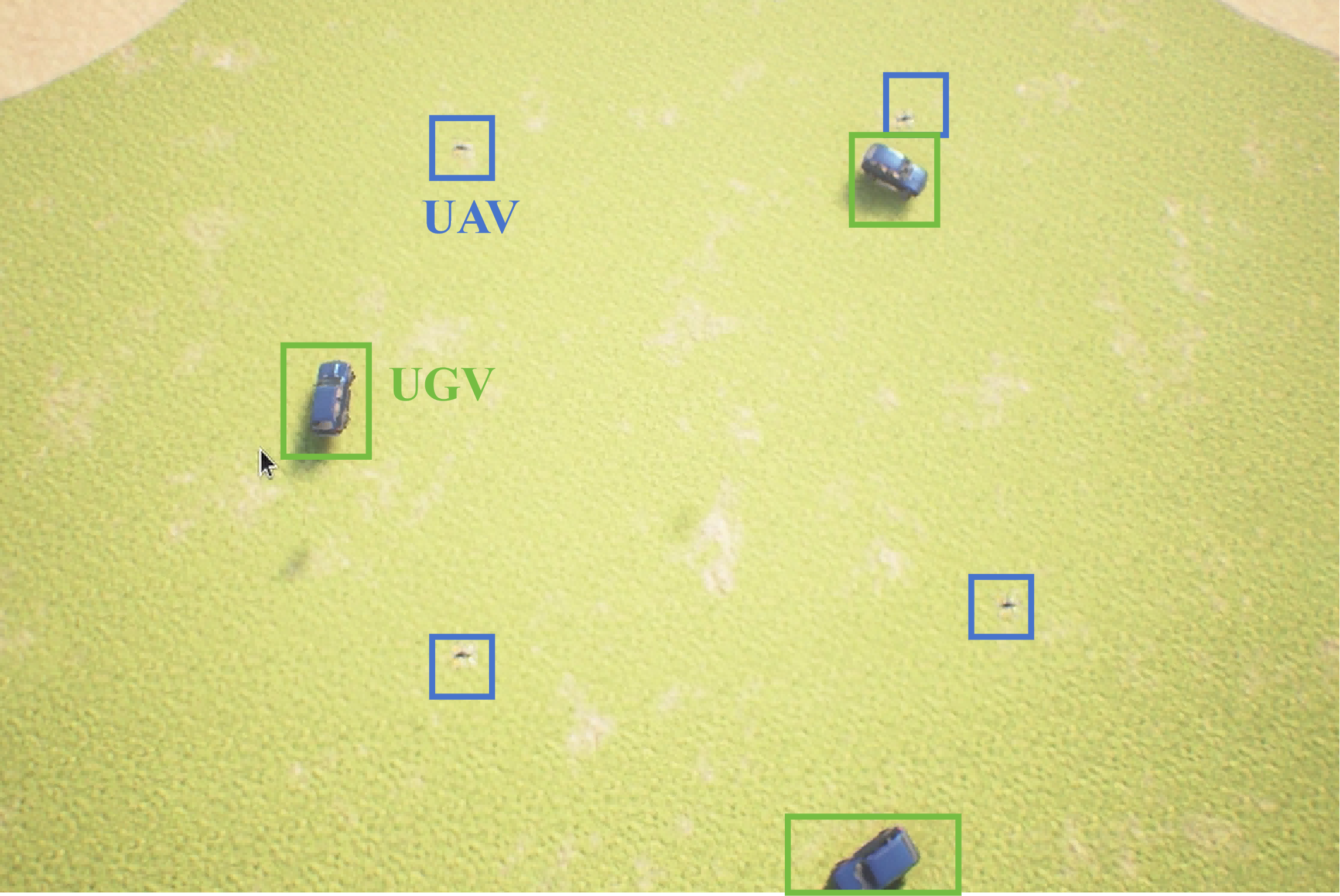}
    \caption{Multi-agent formation.Three UGVs on ground and four UAVs in air executing coordinated trajectories simultaneously.}
    \label{fig:task_multi}
\end{figure}

\begin{table}[tbp!]
\centering
\caption{Performance metrics during the multi-agent simulation experiment with UE "NoDisplay" mode.}
\begin{tabular}{cc}
\hline
\textbf{Metric} & \textbf{Performance} \\ 
\hline
Frequency of Odometry ROS Topic (Hz) & 25 \\
Frequency of Image ROS Topic (Hz) & 5 \\
FPS of Unreal Engine Simulation & 45–60 \\
Memory Usage of Unreal Engine (MB) & 14,852 \\ \hline
\end{tabular}
\label{tb:task_multi}
\end{table}

\subsection{Multi-agent Formation}
\label{subsec:formation}
To further evaluate the scalability of the AirSimAG platform beyond single UAV–UGV cooperation tasks, a multi-agent experiment was conducted involving three UGVs and four UAVs, as illustrated in Fig.~\ref{fig:task_multi}. In this scenario, the UGVs executed circular trajectories on the ground, while the UAVs simultaneously performed square flight patterns in the air.
The experiment was performed on a workstation equipped with an NVIDIA RTX 4090 GPU (24,564 MiB), an Intel Core i7-14700KF CPU, and 62 GB of system memory. Performance metrics are summarized in Table~\ref{tb:task_multi}. Odometry ROS topics were maintained at frequencies exceeding 25 Hz, while image ROS topics operated above 5 Hz.
These results confirm that AirSimAG can support simultaneous operation of multiple heterogeneous agents without compromising simulation fidelity. The platform successfully handled complex multi-agent behaviors, demonstrating its scalability and suitability as a testbed for evaluating coordinated control strategies, perception-driven decision making, and multi-agent cooperation under realistic sensing and computational constraints.

\section{Conclusion}
\label{sec:conclusion}
This paper introduces AirSimAG, a simulation platform for heterogeneous air–ground robotic systems. The platform provides a decoupled architecture that enables independent vehicle simulation APIs, multi-agent communication, and synchronized multi-sensor data streams for UAV and UGV platforms. Representative cooperative tasks, including cooperative mapping, air-assisted navigation, cooperative target tracking, and multi-agent coordination, were implemented to illustrate its functionality. The results demonstrate that AirSimAG provides a practical testbed for air–ground collaborative simulation. Future work will leverage this platform to conduct further air–ground cooperative simulations and task experiments.

\section*{Acknowledgments}
Thanks to the support from ZEX Future Technology Co., Ltd. 

\bibliographystyle{IEEEtran}
\bibliography{ref}

\vfill

\end{document}